\documentclass{article}
\usepackage{ijcai19}

\usepackage{times}
\usepackage{soul}
\usepackage{url}
\usepackage[hidelinks]{hyperref}
\usepackage[utf8]{inputenc}
\usepackage[small]{caption}
\usepackage{graphicx}
\usepackage{amsmath}
\usepackage{booktabs}
\usepackage{algorithm}
\usepackage{algorithmic}
\usepackage{amsfonts}
\usepackage{multicol}

\usepackage[colorinlistoftodos,textwidth=4cm,shadow]{todonotes}

\newcounter{Oleksii}

\newcounter{Sergey}

\title{Catalyst.RL: A Distributed Framework for Reproducible RL Research}

\author{
Sergey Kolesnikov\thanks{equal contribution}$^{13}$
\And
Oleksii Hrinchuk\footnotemark[1]$^{23}$
\affiliations
$^1$Dbrain\\
$^2$Skolkovo Institute of Science and Technology\\
$^3$Moscow Institute of Physics and Technology
\emails
scitator@gmail.com,
oleksii.hrinchuk@gmail.com
}

\begin{document}

\maketitle

\begin{abstract}

Despite the recent progress in deep reinforcement learning field (RL), and, arguably because of it, a large body of work remains to be done in reproducing and carefully comparing different RL algorithms.
We present  \texttt{catalyst.RL}, an open source framework for RL research with a focus on reproducibility and flexibility. 
Main features of our library include large-scale asynchronous distributed training, easy-to-use configuration files with the complete list of hyperparameters for the particular experiments, efficient implementations of various RL algorithms and auxiliary tricks, such as frame stacking, $n$-step returns, value distributions, etc. 
To vindicate the usefulness of our framework, we evaluate it on a range of benchmarks in a  continuous control, as well as on the task of developing a controller to enable a physiologically-based human model with a prosthetic leg to walk and run. 
The latter task was introduced at NeurIPS 2018 AI for Prosthetics Challenge, where our team took the $3$rd place, capitalizing on the ability of \texttt{catalyst.RL} to train high-quality and sample-efficient RL agents.

\end{abstract}

\section{Introduction}\label{sec:intro}

Within the last five years, a huge breakthrough has been made in deep reinforcement learning (RL). Autonomous agents, trained (almost) without any prior information about the particular games first matched and then surpassed professional human players in Atari $2600$~\cite{mnih2015human}, Chess~\cite{silver2017mastering2}, Go~\cite{silver2017mastering}, Dota~$2$~\cite{openaifive2018}, Starcraft II~\cite{deepmind2019}. Deep RL has also been successfully applied to neural networks architecture search~\cite{zoph2016neural} and neural machine translation~\cite{bahdanau2016actor}.

Despite these successes, all modern deep RL algorithms suffer from high variance and extreme sensitivity to hyperparameters~\cite{islam2017reproducibility,henderson2018deep} that makes it difficult to compare algorithms with each other, as well as to evaluate new algorithmic contributions consistently and impartially. For example, in ~\cite{fujimoto2018addressing} it has been shown that small change of hyperparameters in DDPG~\cite{lillicrap2015continuous}, a popular baseline algorithm for continuous control, leads to a significant boost in its performance, which questions the validity of reported performance gain achieved over it in a number of subsequent works.

To make sure that the progress in deep RL research is evaluated consistently and robustly, we need a framework which combines best practices acquired within the last five years and allows a fair comparison between them. The design of such a framework is the problem we attack in this work.

We present \texttt{catalyst.RL}, an open source \texttt{PyTorch} library for RL research written in \texttt{Python}, which offers a variety of tools for the development, evaluation, and fair comparison of deep RL algorithms together with high-quality implementations of modern deep RL algorithms for continuous control. Key features of our library include:

\begin{itemize}
    \item Distributed training with fast communication between various components of the learning pipeline and the support of large-scale setup (e.g., training on a cluster of multiple machines).
    \item Easy to use \texttt{yaml} configuration files with the complete list of hyperparameters for the particular experiments such as neural network architectures, optimizer schedulers, exploration, and gradient clipping schemes.
    \item The ability to run multiple instances of off-policy learning algorithms simultaneously on a shared experience replay buffer to exclude differences caused by exploration and compare different algorithms (architectures, hyperparameters) apples-to-apples.
\end{itemize}

To demonstrate the effectiveness of \texttt{catalyst.RL}, we run a series of extensive computational experiments and benchmark popular RL algorithms for continuous control. More specifically, we compare the performance of DDPG~\cite{lillicrap2015continuous}, TD3~\cite{fujimoto2018addressing}, SAC~\cite{haarnoja2018soft}, and their combinations with distributional value function approximation~\cite{bellemare2017distributional,dabney2018distributional} on a number of environments from OpenAI gym and Deepmind control suite. We also report the results of applying \texttt{catalyst.RL} to the NeurIPS 2018 AI for Prosthetics Challenge task of developing a controller to enable physiologically-based human model with a prosthetic leg to walk and run. This task is especially demanding to the sample-efficiency of the training pipeline because of the extremely slow simulator ($\sim1$ frame per second).
\section{Related work}\label{sec:related}

\paragraph{Reproducibility} In recent years, a lot of concerns about reproducibility of deep reinforcement learning (RL) algorithms have been raised. \cite{islam2017reproducibility} investigated recent advances in policy gradient algorithms for continuous control and came to the conclusion that claimed state-of-the-art results are difficult to reproduce due to general variance in algorithms and extreme sensitivity to hyper-parameter tuning. Later, this analysis was extended by \cite{henderson2018deep}, which showed that entirely different levels of performance might be achieved by the same algorithms if taken from different codebases. They also investigated common flaws in reporting evaluation metrics and provided recommendations on their improvement. \cite{machado2018revisiting} analyzed the diversity of evaluation methodologies in Arcade Learning Environment and highlighted some key concerns and best practices for reporting the experimental results. Recently, \cite{colas2018many} has proposed to look at reproducibility in deep RL through the lens of statistical hypothesis testing and \cite{nagarajan2018deterministic} has investigated the effects of different sources of non-determinism on the performance of deep Q-learning algorithm.

\paragraph{Codebases} In order to fight reproducibility crisis and make recent advances in deep RL more accessible to the wider audience, a large number of codebases have been released. \cite{duan2016benchmarking} conducted an extensive comparison of popular deep RL algorithms for continuous control implemented in \textit{rllab}\footnote{\url{https://github.com/rll/rllab}} framework. OpenAI presented \texttt{Baselines}\footnote{\url{https://github.com/openai/baselines}}, reference implementations of various RL algorithms, and \texttt{Spinning Up}\footnote{\url{https://github.com/openai/spinningup}}, an educational resource aimed to lower the entry threshold into RL research. 
\cite{liang2018rllib} introduced \texttt{RLLib}\footnote{\url{https://github.com/ray-project/ray/tree/master/python/ray/rllib}}, an open-source library for RL that offered both a collection of reference algorithms and scalable primitives for composing new ones. \cite{castro2018dopamine} presented \texttt{Dopamine}\footnote{\url{https://github.com/google/dopamine}}, a research framework for fast prototyping of reinforcement learning algorithms in \texttt{TensorFlow}. \cite{schaarschmidt2018rlgraph} released \texttt{RLGraph}, a library for designing and executing high performance RL computation graphs in both static graph and define-by-run paradigms. 
Recently, \cite{gauci2018horizon} has introduced \texttt{Horizon}\footnote{\url{https://github.com/facebookresearch/Horizon}}, Facebook's open source applied RL platform, and \cite{fan2018surreal} presented \texttt{Surreal}\footnote{\url{https://github.com/SurrealAI/surreal}}, a distributed RL framework specialized for applications in robotics.
\section{Background}\label{sec:background}

In this section, we introduce necessary notation and briefly revise RL algorithms for continuous control we use in our experiments.

\paragraph{RL problem statement} In reinforcement learning, the agent interacts with the environment modeled as a Markov Decision Process (MDP). MDP is a five-tuple $\left(\mathcal{S}, \mathcal{A}, r, P, \gamma \right)$ where $\mathcal{S}$ and $\mathcal{A}$ are the state and action spaces, $r:\mathcal{S}\times\mathcal{A}\rightarrow\mathbb{R}$ is the reward function, $P:\mathcal{S}\times\mathcal{A}\times\mathcal{S}\rightarrow\mathbb{R}_{\geq 0}$ denotes probability density of transitioning to the next state $\mathbf{s}_{t+1} \in \mathcal{S}$ from the current state $\mathbf{s}_{t} \in \mathcal{S}$ after executing action $\mathbf{a}_t \in \mathcal{A}$, and $\gamma \in [0,1]$ is the discount factor. The goal of RL is to maximize the expected return.

In our formulation the action space is continuous $\mathcal{A}\subseteq\mathbb{R}^d$ and the agent decides what action to take according to either deterministic $\mu_\theta:\mathcal{S}\rightarrow\mathcal{A}$ or stochastic $\pi_\theta:\mathcal{S}\rightarrow\left(\mathcal{A}\rightarrow\mathbb{R}_{\geq 0}\right)$ policy represented as neural net with parameters $\theta$. $Z_\phi(\mathbf{s}_t,\mathbf{a}_t) \approx \sum_{t'>=t} \gamma^{t'-t}r(\mathbf{s}_{t'},\mathbf{a}_{t'})$ stands for the approximation of discounted return and $Q_\phi(\mathbf{s}_t,\mathbf{a}_t) = \mathbb{E}Z_\phi(\mathbf{s}_t,\mathbf{a}_t)$ stands for Q-function approximation.

\paragraph{DDPG} Deep Deterministic Policy Gradient (DDPG) is off-policy reinforcement learning algorithm applicable to continuous action spaces~\cite{lillicrap2015continuous}. It usually consists of two neural nets: one of them (actor) approximates deterministic policy and the other (critic) is used for Q-function approximation. The parameters of these networks are updated by the following rules:
\begin{equation*}
\begin{aligned}
\phi \leftarrow \phi &- \alpha \nabla_{\phi} Q_\phi(\mathbf{s}_t,\mathbf{a}_t) \left( y^\text{DDPG} - Q_{\phi}(\mathbf{s}_t,\mathbf{a}_t) \right), \\
\theta \leftarrow \theta &+ \beta \nabla_{\mathbf{a}_t} Q_\phi(\mathbf{s}_t,\mathbf{a}_t) \nabla_{\theta} \mu_{\theta}(\mathbf{s}_t),
\end{aligned}
\end{equation*}
where $y^\text{DDPG} = r_t + \gamma Q_{\tilde\phi}(\mathbf{s}_{t+1},\mu_\theta(\mathbf{s}_{t+1}))$ are TD-targets and transitions $(\mathbf{s}_t,\mathbf{a}_t,r_t,\mathbf{s}_{t+1})$ are uniformly sampled from the experience replay buffer $\mathcal{D}$~\cite{mnih2015human}. TD-targets make use of a target network $Q_{\tilde\phi}$, the architecture of which copies the critic's architecture, and its weights are updated by slowly tracking the learned critic: 
\[\tilde\phi \leftarrow \tau\phi + (1-\tau)\tilde\phi, \quad \tau \ll 1.
\]

\paragraph{TD3} Twin Delayed Deep Deterministic Policy Gradient (TD3)~\cite{fujimoto2018addressing} is a recent improvement over DDPG which adopts Double Q-learning technique to alleviate overestimation bias in actor-critic methods. The differences between TD3 and DDPG are threefold. Firstly, TD3 uses a pair of critics which provides pessimistic estimates of Q-values in TD-targets
\[
Q(\mathbf{s}_{t+1},\mathbf{a}_{t+1}) = \min_{i=1,2} Q_{\tilde{\phi}_i}(\mathbf{s}_{t+1},\mathbf{a}_{t+1}).
\]
Secondly, TD3 introduces a novel regularization strategy, target policy smoothing, which proposes to fit the value of a small area around the target action
\begin{equation*}
\begin{aligned}
y^\text{TD3}&= r_t + \gamma Q(\mathbf{s}_{t+1},\mathbf{a}_{t+1}), \\
\mathbf{a}_{t+1} &= \mu_\theta(\mathbf{s}_{t+1}) + \boldsymbol{\epsilon},\quad\boldsymbol{\epsilon}\sim \text{clip}(\mathcal{N}(0,\sigma I),-c,c).
\end{aligned}
\end{equation*}
Thirdly, TD3 updates an actor network less frequently than a critic network (typically, one actor update for two critic updates). 

In our experiments, the application of the first two modifications led to much more stable and robust learning. 
Updating the actor less often did not result in better performance, thus, that modification was omitted in the experiments. However, our implementation allows to include it back.

\paragraph{SAC} Soft Actor-Critic (SAC)~\cite{haarnoja2018soft} optimizes more general maximum entropy objective and learns a stochastic policy $\pi_\theta(\mathbf{a}_t|\mathbf{s}_t)$, usually parametrized in a way to support reparametrization trick (e.g., Gaussian policy with learnable mean and variance or policy represented via normalizing flow) and a pair of Q-functions. The parameters of all networks are updated by the following rules:

\begin{equation*}
\begin{aligned}
\phi_i &\leftarrow \phi_i - \alpha \nabla_{\phi_i} Q_{\phi_i}(\mathbf{s}_t,\mathbf{a}_t) \left( y^\text{SAC} - Q_{\phi_i}(\mathbf{s}_t,\mathbf{a}_t) \right), \\
\theta &\leftarrow \theta + \beta \nabla_{\theta} \left( Q_{\phi_1}(\mathbf{s}_t,\mathbf{a}_t) - \log\pi_\theta(\mathbf{a}_t|\mathbf{s}_t) \right), \\
y^\text{SAC} &= r_t + \gamma ( \min_{i=1,2} Q_{\tilde{\phi}_i}(\mathbf{s}_{t+1},\mathbf{a}_{t+1}) -\log\pi_\theta(\mathbf{a}_{t+1}|\mathbf{s}_{t+1})).
\end{aligned}
\end{equation*}

\paragraph{Distributional RL} A distributional perspective on RL suggests to approximate the distribution of discounted return instead of the approximation its first moment (Q-function) only. In this work, we analyze two different ways to parametrize this distribution (also referred to as \textit{value distribution}), namely categorical~\cite{bellemare2017distributional} and quantile~\cite{dabney2018distributional}.

Categorical return approximation represents discounted return as $Z_\phi(\mathbf{s}_t,\mathbf{a}_t) = \sum_{i=1}^N p_{\phi,i}(\mathbf{s}_t,\mathbf{a}_t) \delta_{z_i}$, where $\delta_z$ denotes a Dirac at $z \in \mathbb{R}$ and atoms $z_i$ split the distribution support $[V_\text{MIN},V_\text{MAX}] \subset \mathbb{R}$ by $N-1$ equal parts. Its parameters are updated by minimizing the cross-entropy between $Z_\phi(\mathbf{s}_t,\mathbf{a}_t)$ and $\Phi\tilde{\mathcal{T}} Z_{\tilde{\phi}}(\mathbf{s}_t,\mathbf{a}_t)$ --- the Cramer projection of target value distribution $\tilde{\mathcal{T}} Z_{\tilde{\phi}}(\mathbf{s}_t,\mathbf{a}_t) = r_t + \gamma Z_{\tilde{\phi}}(\mathbf{s}_{t+1},\mathbf{a}_{t+1})$ onto the distribution support.

Quantile approximation sticks to the ``transposed" parametrization $Z_\phi(\mathbf{s}_t,\mathbf{a}_t) = \frac{1}{N}\sum_{i=1}^N \delta_{z_{\phi, i}(\mathbf{s}_t,\mathbf{a}_t)}$, which assigns equal probability masses $\frac{1}{N}$ to the set of learnable atom positions $\{z_{\phi,i}(\mathbf{s}_t,\mathbf{a}_t)\}_{i=1}^N$. Distribution parameters are updated by minimizing the quantile regression Huber loss. The exact formulas are cumbersome and omitted here for brevity, thus we refer the interested reader to~\cite{dabney2018distributional}.
\section{Framework}\label{sec:framework}

\begin{figure}
\centering
\includegraphics[width=0.5\textwidth]{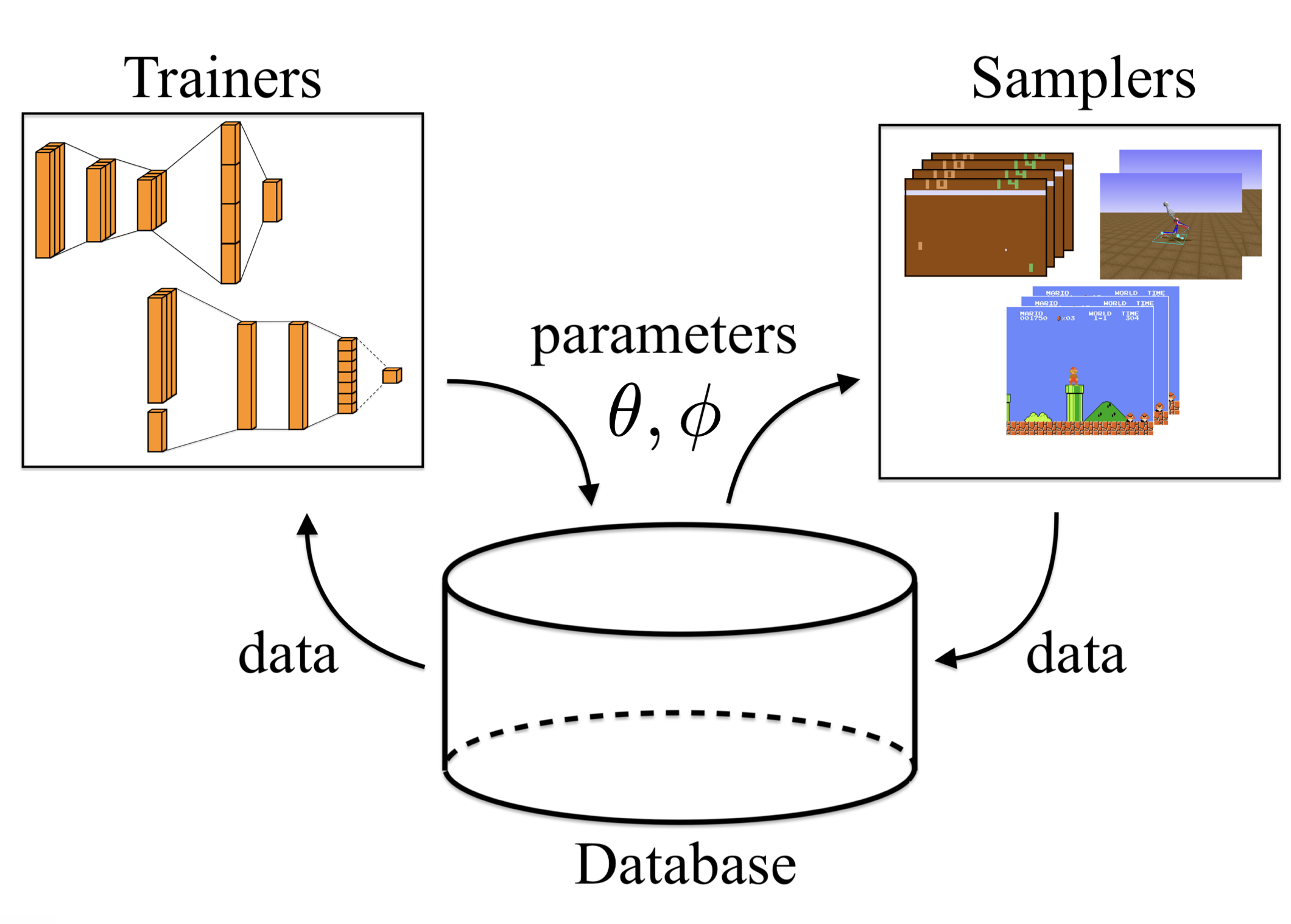}
\caption{\texttt{catalyst.RL} achitecture. Samplers interact with the environment and gather training data. Trainers retrieve collected data and update parameters of value function and policy approximators. All communication is conducted through redis database.}
\label{fig:rl_arena}
\end{figure}

In this section, we introduce \texttt{catalyst.RL} and discuss the key design decisions which make it a suitable framework for flexible and reproducible deep RL research.

\paragraph{Flexibility} We believe that flexibility is a crucial aspect in any codebase suitable for RL research. The code should be readable and well-structured to minimize the efforts of the researcher who wants to adapt it to one's needs. 
\texttt{catalyst.RL} is developed in strict conformity with high-quality code standards that make it easy to change available algorithms or develop the new ones.

\paragraph{Scalability} Another important feature of our library is scalability. \texttt{catalyst.RL} can be easily used on a laptop, a single server or a huge cluster, which consists of multiple machines, without any code overhead capitalizing on its architectural design. More specifically, our solution is based on interconnected nodes of three types, namely \textit{trainers}, \textit{samplers}, and \textit{database}. Trainers are the nodes with GPUs which constantly update the parameters of neural networks representing policies and value functions. Samplers are CPU-based nodes which obtain policy networks weights and utilize them for inference only to collect training data. The database serves as an intermediary between trainers and samplers: it receives neural networks weights from trainers to send them to samplers; and receives data collected by samplers to send them for training. 
A critical feature of such a system is stability and robustness to individual breakdowns of its components, as all nodes are entirely independent and interchangeable.

\paragraph{Compositionality} \texttt{catalyst.RL} provides a onvenient API for the development and evaluation of RL algorithms. In addition to the described above trainers, samplers, and database, which implement all necessary low-level operations for fast and efficient data transmission, it contains two additional abstract classes --- \textit{algorithm} and \textit{agent}. Algorithm implements the computation graphs of RL algorithms. Agent implements the neural network architectures used for policy (actor) and value function (critic) approximation.

Such compositionality allows for fast prototyping and easy debugging of RL pipelines, and usually, only a small part of the whole system needs to be changed. For example, to implement SAC with policies represented as normalizing flows~\cite{haarnoja2018latent}, we only need to add one new layer (coupling layer) and use it for the construction of a novel actor.

\paragraph{Reproducibility} To ensure reproducibility and interpretability of the results obtained by \texttt{catalyst.RL}, we predefine the same set of validation seeds for all models tested and compare them on this set only. To eliminate the evaluation bias coming from distorting actions produced by the policies with various exploration techniques, we run several samplers in deterministic mode and report their performance on the learning curves.

To keep track of various important metrics and visualize them during training we utilize the TensorboardX plugin. By default we log average reward, actor and critic losses, a number of data samples/parameters updates per second and any additional metrics of interest can be easily added by the user. We also provide a functional framework for saving the source code of each particular experiment, to exclude the situation when seemingly the same hyperparameter setup produces entirely different results due to untracked experimental changes in the code, but not due to the variance of the algorithm.

\section{Experiments}\label{sec:experiments}

\subsection{Continuous control benchmarks}

We evaluate the implementations of off-policy deep RL algorithms for continuous control in \texttt{catalyst.RL} on a variety of benchmarks. More specifically, we run distributed versions of DDPG~\cite{lillicrap2015continuous}, TD3~\cite{fujimoto2018addressing}, SAC~\cite{haarnoja2018soft}, and their combinations with distributional RL, both categorical~\cite{bellemare2017distributional} and quantile~\cite{dabney2018distributional}, on three environments from OpenAI gym\footnote{\url{https://gym.openai.com}} (LunarLander, BipedalWalker-v2, BipedalWalkerHardcore-v2) and four environments from DeepMind control suite (Reacher-hard, Cheetah-run, Walker-run, Humanoid-walk).

In the absence of substantial computational resources required for a more thorough search, we decided to stick to them. Specifically, we use $2$-layer fully connected networks with ReLU nonlinearity and $[400, 300]$ neurons in hidden layers for both actor and critic; critic receives action as input to the second layer; both network parameters are updated using Adam with a learning rate of $10^{-4}$; layer normalization~\cite{ba2016layer} is employed between layers.

For value distribution approximation (either categorical or quantile) we used $101$ atoms which showed better results in preliminary experiments comparing to a more common choice of $51$ atoms. Note, that increasing the resolution of the distribution does not incur an increased computational cost due to the efficient vectorized implementations of distributional losses. The support of categorical distribution was chosen to be $[-100;100]$ to capture the maximum possible discounted return for $\gamma=0.99$ in DeepMind control suite environments. Finally, the reward scale hyperparameter in SAC was set to $150$ as a result of a grid search over $[50, 100, 125, 150, 175, 200]$ on Walker-run environment.

For the complete lists of hyperparameters represented as \texttt{yaml} configuration files necessary to reproduce the results of our experiments, we refer the reader to our repository\footnote{\url{https://github.com/catalyst-team/catalyst}}.

\subsubsection{Algorithms comparison}

In our experiments, we investigated how all $3$ algorithms mentioned above compared to each other and how two improvements, namely $n$-step return and distributional RL, affected their performance. Concretely, we considered two values of $n \in \{1, 5\}$ and three modes for distributional RL: no distribution, categorical distribution\footnote{The combination of distributed DDPG with categorical distribution is also known as D4PG in the literature.}, and quantile distribution, which resulted in $6$ different configurations of each learning algorithm. Each training run used $20$ CPU samplers for collecting experience, $4$ CPU samplers for evaluating the algorithm (ran in inference mode with no exploration noise), and $1$ GPU for training. All experiments were terminated after $12$ hours of training.

Table~\ref{tab:aggregated_score} shows the aggregated scores of each algorithm configuration, computed by adopting the evaluation technique from~\cite{schulman2017proximal}. Specifically, each algorithm was run on all $6$ environments, and the average total reward of the last $100$ episodes was computed. Then, we shifted and scaled scores of each environment to be in $[0,1]$ and averaged them to obtain a single scalar score for each algorithm configuration.

\begin{table}[htb!]
\caption{Aggregated score (from $0$ to $1$) of each tested configuration, calculated as the mean return over last $100$ episodes scaled to $[0,1]$ and averaged over all $6$ environments.}
\label{tab:aggregated_score}
\begin{center}
  \begin{tabular}{lcccccc}
    \toprule
    $Z_\phi(\mathbf{s},\mathbf{a})$ & \multicolumn{2}{c}{None} & \multicolumn{2}{c}{Categorical} & \multicolumn{2}{c}{Quantile}\\
    $n$-step & 1 & 5 & 1 & 5 & 1 & 5 \\
    \midrule
    DDPG & 0.56 & 0.60 & \textbf{0.64} & 0.61 & \textbf{0.67} & 0.50 \\
    TD3 & 0.48 & 0.52 & 0.54 & \textbf{0.66} & 0.38 & 0.60 \\
    SAC & 0.46 & 0.52 & 0.52 & 0.45 & 0.32 & 0.48 \\
    \bottomrule
  \end{tabular}
\end{center}
\end{table}

\begin{figure*}
\centering
\includegraphics[width=\textwidth,height=\textheight,keepaspectratio]{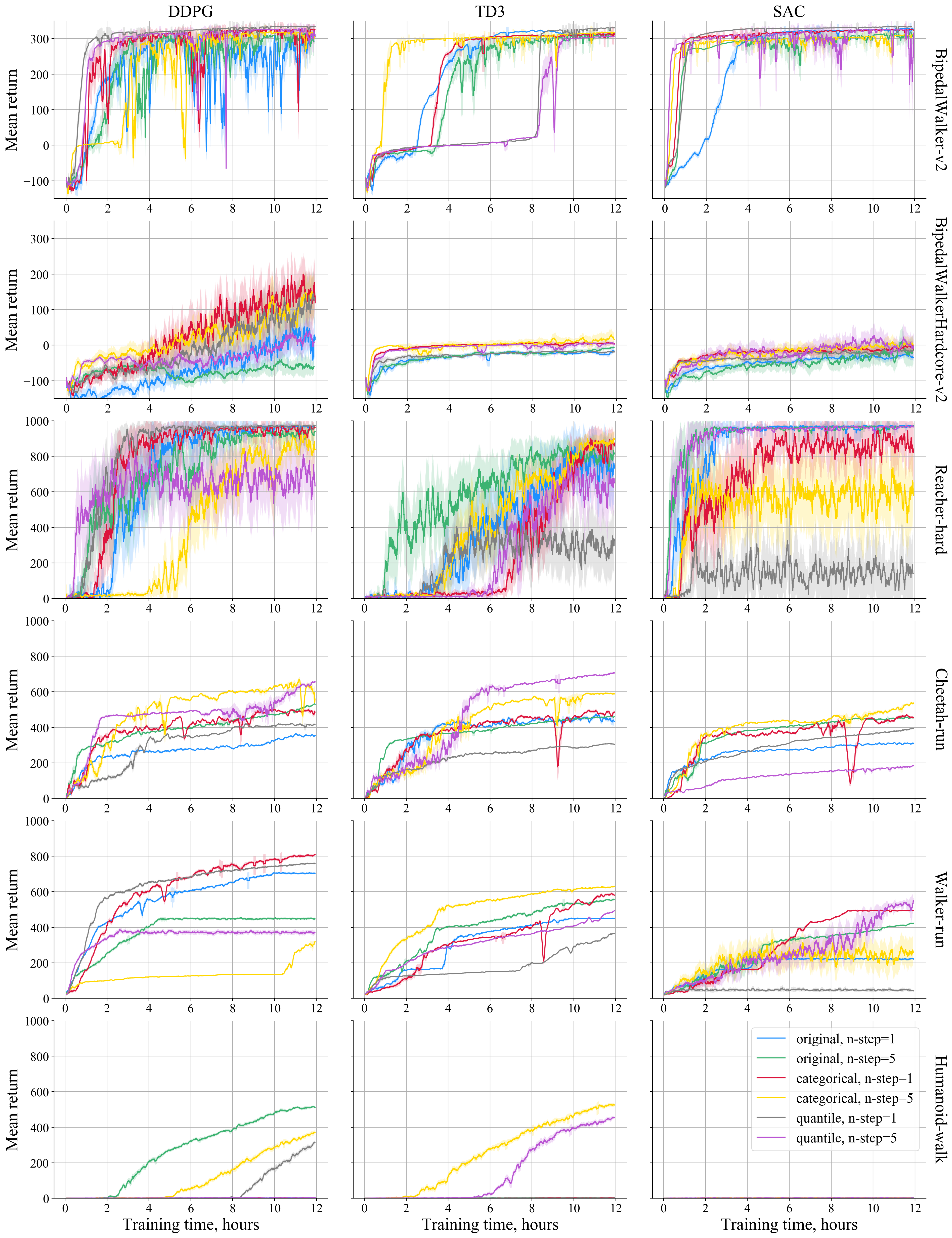}
\caption{Performance of DDPG, TD3 and SAC on $6$ benchmark tasks with various $n$-step returns and value distribution approximations. Shaded regions represent half of the standard deviation in return.}
\label{fig:results1}
\end{figure*}

Figure~\ref{fig:results1} shows the results of our experiments. Surprisingly, it is difficult to name an algorithm which consistently outperforms all the others. However, there are some insights revealed by our experiments.

\begin{enumerate}
    \item BipedalWalker-v2 and Reacher-hard are relatively easy tasks as all the algorithms solve them eventually, at least in one of the tested configurations.
    \item BipedalWalkerHardcore-v2 and Humanoid-walk are difficult tasks as only a fraction of $18$ tested algorithmic configurations managed to find a policy with nonzero reward. Interestingly, DDPG which seems to be the least competitive algorithm clearly outperforms TD3 and SAC on BipedalWalkerHardcore-v2. We hypothesize that this is because DDPG has a higher variance and, thus, is more exploratory, while its competitors are more robust and stable, which make them quickly converge to a safe yet suboptimal solution.
    \item Cheetah-run and Walker-run represent the middle ground environments, on which all algorithms manage to achieve reasonable levels of performance.
    \item In most cases, $n$-step return or distributional RL improves the performance of baseline algorithms, yet not always. More rigorous and computationally expensive evaluation is required to understand in which situations adding them is beneficial.
\end{enumerate}

To sum up, our experiments suggest that there is no ``silver bullet" which can be successfully applied to any environment. Each specific case requires rigorous algorithm and hyperparameter selection process.

\subsection{AI for Prosthetics Challenge}

\begin{figure}[ht]
\centering
\includegraphics[width=0.48\textwidth]{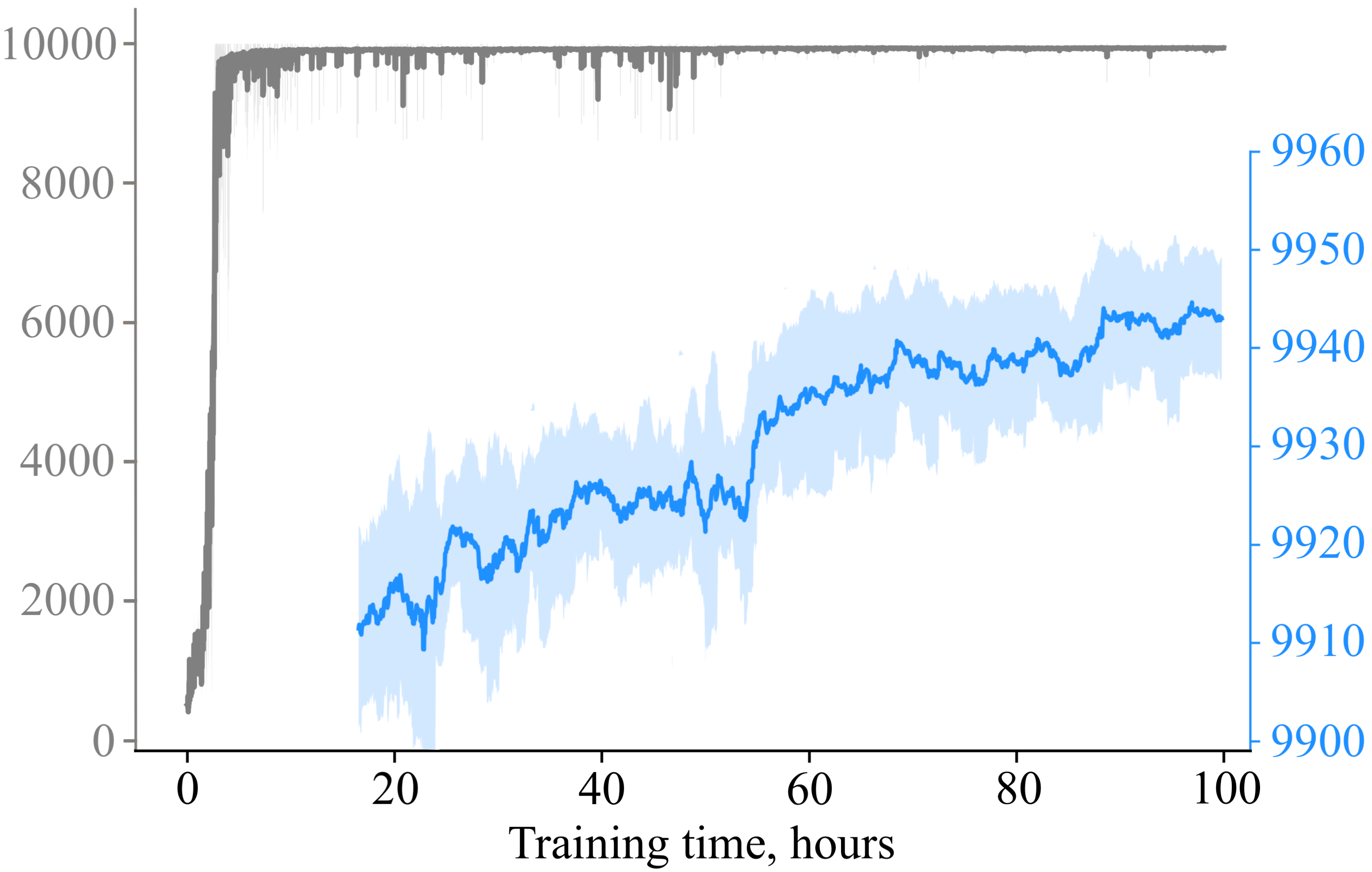}
\caption{AI for Prosthetics Challenge. Average reward of RL agent (TD3 with quantile value distribution approximation) trained with \texttt{catalyst.RL} (gray) and its rescaled version (blue) to better visualize the progress made after the first $20$ hours of training. Shaded region represents the standard deviation of the reward.}
\label{fig:l2run}
\end{figure}

We also applied \texttt{catalyst.RL} to the task of developing a controller to enable a physiologically-based human model with a prosthetic leg to walk and run. 
This task was introduced at NeurIPS 2018 AI for Prosthetics Challenge\footnote{\url{https://www.crowdai.org/challenges/neurips-2018-ai-for-prosthetics-challenge}}, where our team took the $3$rd place, capitalizing on the sample-efficient off-policy algorithms implementations from \texttt{catalyst.RL}. In this problem, the sample-efficiency was especially important as the simulator generating data was extremely slow ($\sim1$ observation per second).

The task was to build a controller which mapped states $\mathbf{s}_t \in \mathbb{R}^{344}$, that consist of current positions, velocities, accelerations of joints and body parts, muscles activities to muscle activations $\mathbf{a}_t \in \mathbb{R}^{19}$. The agent had to follow a requested velocity vector which changed over time.

The reward is defined as:
\[
r(\mathbf{s}_t,\mathbf{a}_t) = 10 - \left(\mathbf{v}_{t} - \mathbf{u}_{t}\right)^2 -0.001 \cdot \|\mathbf{a}_t\|_2,
\]
where $\mathbf{v}_t$ is the agent velocity, $\mathbf{u}_t$ is the target velocity vector we need to follow, and the last term is a penalty muscle term to encourage more human-like gait. The episode ends after $1000$ time steps or if the pelvis of the agent falls below $0.6$ meters, resulting in a maximum possible reward of $10000$.

The ability of \texttt{catalyst.RL} to run multiple different algorithms and/or hyperparameter setups in parallel allowed us to quickly select the best candidate for the final solution -- the combination of TD3 algorithm with quantile distribution approximation. Figure~\ref{fig:l2run} depicts the learning curve (average return) of our agent trained with $24$ parallel CPU samplers on $1$ Nvidia $1080$ Ti GPU.
\section{Conclusion}\label{sec:conclusion}

In this paper, we introduced \texttt{catalyst.RL}, an open source library for RL research, which offers a variety of tools for the development and evaluation of (presently, off-policy) deep RL algorithms.

To the best of our knowledge, the results presented here for DDPG, SAC and TD3, were obtained with well-performing distributed implementations of these algorithms, however, the learning curves are based on the default, most likely, highly suboptimal set of hyperparameters. Thus, we expect that it may be possible to improve significantly on the reported results, especially if tuning hyperparameters for each particular task.

We hope that our framework will be useful to both professional researchers and RL enthusiasts, and will help to overcome the current reproducibility crisis. We are looking forward to exciting advances in RL research powered by \texttt{catalyst.RL} and heartily welcome contributions from the broader computer science community to make it better.

\paragraph{Future work} Several elements are missing from the current release of \texttt{catalyst.RL}, such as on-policy algorithms and off-policy algorithms for discrete control. In the future, we plan to continue the development and support of \texttt{catalyst.RL} by adding new features and algorithms, as well as to heavily benchmark them to keep track of the state-of-the-art results and get a better understanding of the applicability of certain methods in deep reinforcement learning.

\bibliographystyle{named}
\bibliography{ijcai19}

\end{document}